\documentclass{llncs}

\usepackage{makeidx} 

\usepackage[misc]{ifsym}

\usepackage{graphicx}
\usepackage{float}
\usepackage{subfig}
\graphicspath{{./figures/}}

\usepackage{amsmath}
\usepackage{amsfonts}
\usepackage{amssymb}
\usepackage{bm}
\usepackage{mathtools}

\usepackage{tikz}
\usetikzlibrary{bayesnet}

\usepackage{minitoc} 
\usepackage[toc,page]{appendix} 
\usepackage[font=small,labelfont=bf]{caption} 
\usepackage{soul}
\usepackage{esvect}
\usepackage{algorithm}
\usepackage{algorithmic}
\usepackage{pgfgantt}
\usepackage{hyperref}

\DeclareMathOperator{\transp}{\mathrm{T}}
\newcommand{\norm}[1]{\left\lVert#1\right\rVert} 
\DeclareMathOperator*{\argmin}{\arg\!\min} 

\usepackage{widetable}
\usepackage{booktabs}

\begin{document}

\frontmatter          

\pagestyle{headings}  

\mainmatter              

\title{MRI Super-Resolution using Multi-Channel Total Variation}

\titlerunning{MRI Super-Resolution Using a Multi-Channel Total Variation Prior}  

\author{Mikael Brudfors\inst{1} (\Letter) \and Ya\"{e}l Balbastre\inst{1} \and Parashkev Nachev\inst{2} \and John Ashburner\inst{1}}

\authorrunning{Mikael Brudfors et al.} 

\institute{The Wellcome Centre for Human Neuroimaging, UCL, London, UK\\
	\email{\{mikael.brudfors.15,y.balbastre,j.ashburner\}@ucl.ac.uk}
	\and
	UCL Institute of Neurology, London, UK\\
	\email{p.nachev@ucl.ac.uk}}

\maketitle              

\begin{abstract}
This paper presents a generative model for super-resolution in routine 
clinical magnetic resonance images (MRI), of arbitrary orientation and 
contrast. The model recasts the recovery of high resolution images as 
an inverse problem, in which a forward model simulates the slice-select 
profile of the MR scanner. The paper introduces a prior based on 
multi-channel total variation for MRI super-resolution. Bias-variance 
trade-off is handled by estimating hyper-parameters from the low 
resolution input scans. The model was validated on a large database of 
brain images. The validation showed that the model can improve brain 
segmentation, that it can recover anatomical information between images 
of different MR contrasts, and that it generalises well to the large 
variability present in MR images of different subjects. The 
implementation is freely available at 
\url{https://github.com/brudfors/spm_superres}.
\end{abstract}


\section{Introduction}
The cost of storing data has decreased dramatically in recent decades and large databases of patient images are now contained within most hospitals. In fact, up until 2010, five billion medical imaging studies had been conducted worldwide \cite{roobottom2010radiation}. However, although large quantities of data exist, much of the information available in that data lies latent, its potential uses yet untapped \cite{Smith2018263}. This is because hospital grade MR data commonly are of lower quality than images collected in a research context. This quality difference hampers automated neuroimaging analysis and originates from the scans being acquired in shorter times, leading to greater prevalence of different types of artefacts, in particular thick-sliced data.

The anatomical information in hospital MR images differs from images used in research studies in being distributed across a larger number of low resolution (LR) scans. A collection of MR images from a patient scanning session therefore commonly contains: (1) thick-sliced data and (2) multiple MR contrasts (e.g. T1- and T2-weighted images). For example, in a research study, one might acquire a single, time-consuming, high-resolution (HR) T1-weighted image. Whereas in a hospital study, one might acquire several quick LR scans of varying contrasts and with thick slices. Multi-contrast MR images are acquired in order to highlight different anatomical information, in particular pathology. Hospital MR images can therefore be thought of as information distributed across multiple scans.

If the distributed information across hospital MR images could be consolidated it could be possible to achieve closer to research quality imaging using ordinary, hospital quality data. Reconstructing HR images in such a way would in turn enable large scale studies, which have been difficult to perform in a research environment because of the expense of scanning a large number of patients within a research context. A framework that could merge this distributed information would therefore be of value in, for example, research relying on learning from a population of MR images (see e.g \cite{blaiotta2017generative,havaei2017brain,ronneberger2015u}).

This paper proposes a generative model that given a set of patient MR images, which could be of different contrasts, reconstructs isotropic HR images, using a super-resolution technique. The paper furthermore introduces the multi-channel total variation norm as a prior for super-resolution recovery in MR images. The model additionally does not require any HR reference data, but could utilise such data if available. The incentives for choosing a generative model are: (1) the large variability in clinical MRI, which can be difficult to model in a discriminative setting; and (2), that learning methods potentially hallucinate information not actually present, which should be avoided when accurate patient diagnosis is of interest. Because of the challenge in quantifying the results of quality enhancement applied to MR images, the model was validated on thick-sliced data generated from the IXI dataset\footnote{The IXI dataset contains nearly 600 HR MR images from healthy subjects, of multiple MR contrasts. It is available from \url{http://brain-development.org/ixi-dataset/}.}.

\section{Background}
Super-resolution (SR) methods construct HR images from several observed LR images, where each LR image transforms and samples the same HR scene. SR methods are therefore able to utilise distributed information available in collections of images, such as patient MR data. SR have been studied in MRI for numerous applications \cite{van2012super}, and shown to improve the resolution and signal-to-noise ratio favourably compared with direct HR acquisition \cite{plenge2012super}. 

Most SR work in MRI have focused on images of the brain. This is because SR methods have a high dependency on precise registration of the observed LR images, and for the brain, simply a rigid alignment is sufficient. SR methods applied to brain MR scans can be classified as either single- or multi-modal. Single-modality methods combine LR images of the same MR contrasts into a single HR image (e.g. three T1-weighted scans of the same subject). Numerous methods have been proposed for single-modality SR based on e.g. the acquisition model \cite{greenspan2002mri,poot2010general}, image patches \cite{manjon2010non}, random forests \cite{alexander2014image} and convolutional neural networks \cite{wang2016accelerating}. A multitude of prior terms (or regularisation methods) have also been investigated, e.g. Tikhonov \cite{poot2010general}, total variation \cite{ebner2018volumetric} and Beltrami \cite{odille2015motion}.

Multi-modality methods have so far been less frequently explored than their single-modality counterparts. In these techniques the goal is to utilise anatomical information distributed over a number of various quality images, of different MR contrasts (e.g. two T1- and one T2-weighted scan). Manjon et al. \cite{manjon2010mri} proposed a multi-modal method that used as input HR reference data and a pre-interpolated version of the LR data. HR images were then reconstructed by averaging voxels in the interpolated version of the LR data using the HR data as reference. In a paper by Rousseau \cite{rousseau2010non}, the main idea was to reconstruct a HR image using one LR image and an intermodality prior, which included information from another HR image. An extension to the method proposed in \cite{manjon2010mri} was suggested by Jafari et al. \cite{jafari2014mri}, which resulted in higher interpolation accuracy. This method used a similarity metric based on the assumption that voxels that have equal distances to a strong edge are more likely to belong to the same tissue type. 

\section{Methods}
Compared to the multi-modal SR approaches mentioned in the previous section, the model proposed here does not require any complex learning, nor any HR reference data (which significantly reduces the applicability of any SR technique in clinical practice). Instead, it relies on the definition of a generative stochastic process. This process assumes that each LR image is generated by selecting thick slices, possibly rotated and/or translated, in a HR image, with the addition of i.i.d. Gaussian noise (e.g. two LR T1- and three LR T2-weighted images are assumed observations of one HR T1- and one HR T2-weighted image, respectively). This yields a conditional probability distribution known as the data likelihood. A HR image is also assumed to result from a random process, characterised by a probability distribution known as the prior. In SR, the prior should favour images with large smooth regions and a few sharp edges. Furthermore, because edges should have the same location in all MR contrasts, a multi-channel total-variation (MTV) distribution is used, which promotes modalities with common smoothness profiles \cite{wen2008efficient}. Estimating the HR images given a set of observed  LR images is then cast as an inference problem in this probabilistic model. In order for the model to generalise well, its hyper-parameters are either estimated from the observed data (Gaussian and MTV parameters) or set in a general and consistent way (slice-selection profile).

\begin{figure}
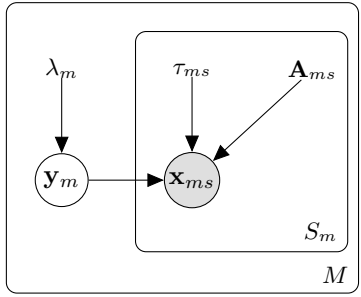

	\centering
	\tikz{
		\node[latent] (y) {${\bf y}_{m}$};
		\node[const, above=of y] (lambda) {$\lambda_m$};
		\node[obs, right=of y] (x) {${\bf x}_{ms}$};
		\node[const, above=of x] (tau) {$\tau_{ms}$};
		\node[const, above=of x, right=of tau] (A) {${\bf A}_{ms}$} ;
		
		\plate[inner sep=0.25cm, xshift=-0.12cm, yshift=0.12cm] {plate1} {(x) (tau) (A)} {$S_m$};
		\plate[inner sep=0.25cm, xshift=-0.12cm, yshift=0.12cm] {plate2} {(y) (lambda) (plate1)} {$M$};	
		
		\edge {lambda} {y};
		\edge {y} {x};
		\edge {tau,A} {x};
	}
	\caption{Graphical model for multi-modality SR in MR scans using a MTV prior. Random variables are in circles. Observed variables are shaded. Plates indicate replication. There are $M$ unknown HR images ${\bf y}_m$. For each HR image there are $S_m$ observed LR images ${\bf x}_{ms}$. Model hyper-parameters are: Gaussian noise of precisions $\tau_{ms}$, projection matrices ${\bf A}_{ms}$, and regularisation parameters $\lambda_m$.}
	\label{fig:improving:gm}
\end{figure}

\subsection{The Generative Model}
Let ${\bf Y} = \{ \{ {\bf y}_m \}_{m=1}^{M} \mid {\bf y}_m \in \mathbb{R}^{N} \}$ denote the unknown HR images of $M$ different MR contrasts and $\bm{\mathcal{X}} = \{ \{ {\bf X}_m \}_{m=1}^{M} \mid {\bf X}_m = \{ {\bf x}_{ms} \}_{s=1}^{S_m} \mid {\bf x}_{ms} \in \mathbb{R}^{N_{ms}} \}$ a set of LR images, where $S_m$ is the number of observed subject images of MR contrast $m$. The joint probability of the model can then be written as:
\begin{align}
p(\bm{\mathcal{X}},{\bf Y}) =
p(\bm{\mathcal{X}} | {\bf Y})p({\bf Y}) = \prod_{m=1}^M \prod_{s=1}^{S_m} p({\bf x}_{ms} | {\bf y}_m) p({\bf Y}).
\label{eq:improving:jp}
\end{align}
A graphical representation of the joint probability in \eqref{eq:improving:jp} is shown in Figure \ref{fig:improving:gm}. 

The conditional distribution of an observed LR image, given an unknown HR image, is assumed to be drawn from a likelihood function that is well-established in the SR literature, the multivariate Gaussian:
\begin{align}
p({\bf x}_{ms} | {\bf y}_m)
& = \mathcal{N} ({\bf x}_{ms} | {\bf A}_{ms} {\bf y}_m,\tau_{ms}^{-1} {\bf I}) \cr
& = \frac{\tau_{ms}^{N_{ms}/2}}{(2 \pi)^{N_{ms}/2}} \exp \left( -\frac{\tau_{ms}}{2} ({\bf A}_{ms}{\bf y}_m - {\bf x}_{ms})^{\transp} ({\bf A}_{ms}{\bf y}_{m} - {\bf x}_{ms}) \right), \cr
\label{eq:improving:cond}
\end{align}
where $\tau_{ms}$ is the precision of the observation noise ($\tau_{ms} = 1/\sigma_{ms}^2$) and ${\bf A}_{ms} \in \mathbb{R}^{N_{ms} \times N}$ a projection matrix. Furthermore, the reconstructions are assumed to all have the same dimensions and 1 mm isotropic voxel size.

The prior probability of the reconstructions is given by:
\begin{align}
p({\bf Y}) = Z^{-1} f ({\bf Y}),
\label{eq:improving:prior}
\end{align}
and is defined by the normalising factor  $Z$, which is independent of ${\bf Y}$, and the MTV prior, defined by:
\begin{align}
f ({\bf Y})&
= \prod_{n=1}^{N} \exp \left( -\sqrt{\sum_{m=1}^M \lambda_{m} \norm{{\bf D}_n {\bf y}_{m}}_2^2} \right),
\label{eq:improving:prior2}
\end{align}
where ${\bf D}_n {\bf y}_{m} \in \mathbb{R}^{3}$ is the finite forward difference at voxel $n$, of reconstruction $m$. Using the MTV norm can intuitively be thought of as a method encouraging patterns where regions of larger gradient magnitude are similar over all MRI contrasts, through its sparsity inducing effect. Note that for $M=1$, the MTV norm reduces to ordinary, isotropic TV. 

\subsection{Model Optimisation}
A maximum a posteriori (MAP) estimate of the HR images can be obtained from the joint probability in \eqref{eq:improving:jp} by minimising its negative logarithm: 
\begin{align}
& \argmin_{\bf Y} \left\{ - \ln  p({\bf Y} | \bm{\mathcal{X}})  \right\},
\label{eq:improving:energy}
\end{align}
where
\begin{align}
- \ln p({\bf Y} | \bm{\mathcal{X}})  
=& \sum_{m=1}^M \left( \sum_{s=1}^{S_m} \left( \frac{\tau_{ms}}{2} \norm{{\bf A}_{ms} {\bf y}_m - {\bf x}_{ms}}_2^2 \right)\right) \cr
& + \sum_{n=1}^{N} \left( \sqrt{\sum_{m=1}^M \lambda_m \norm{{\bf D}_n {\bf y}_{m}}_2^2}\right) + \text{const}.
\end{align}
The optimisation problem  in \eqref{eq:improving:energy} is solved by an alternating direction method of multipliers (ADMM) algorithm \cite{boyd2011distributed}. ADMM is part of a class of optimisation methods called proximal algorithms. ADMM recasts the energy minimisation as a constrained problem, from which an augmented Lagrangian  can be formulated. The Lagrangian is then minimised in an alternating fashion until a convergence criterion is met. An ADMM algorithm was chosen because it is straightforward to implement and gives competitive results when solving TV problems in neuroimaging \cite{dohmatob2014benchmarking}.

\begin{figure*}
	\centering 	
	\subfloat[]{\includegraphics[width=0.31\textwidth]{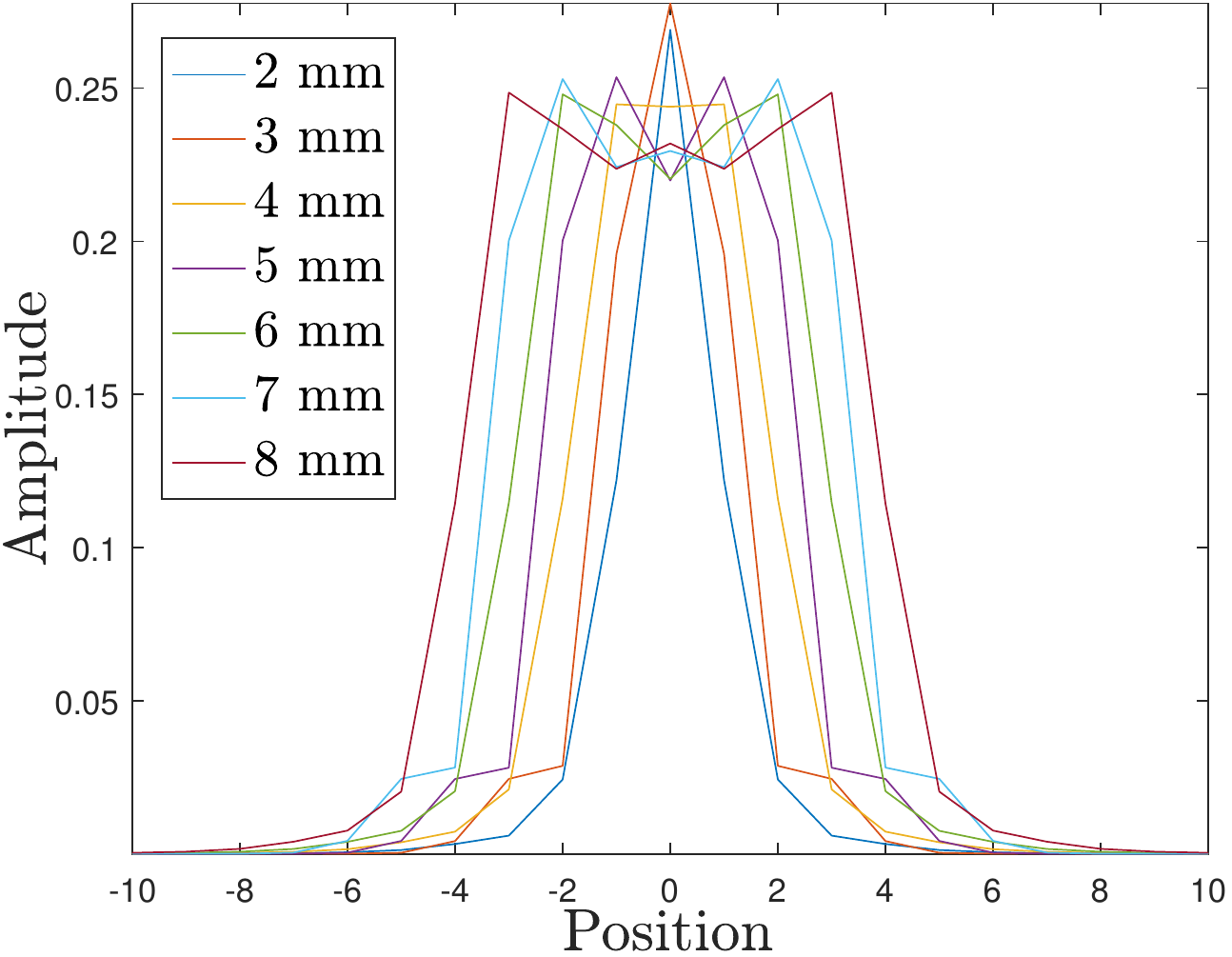}} \quad
	\subfloat[]{\includegraphics[width=0.31\textwidth]{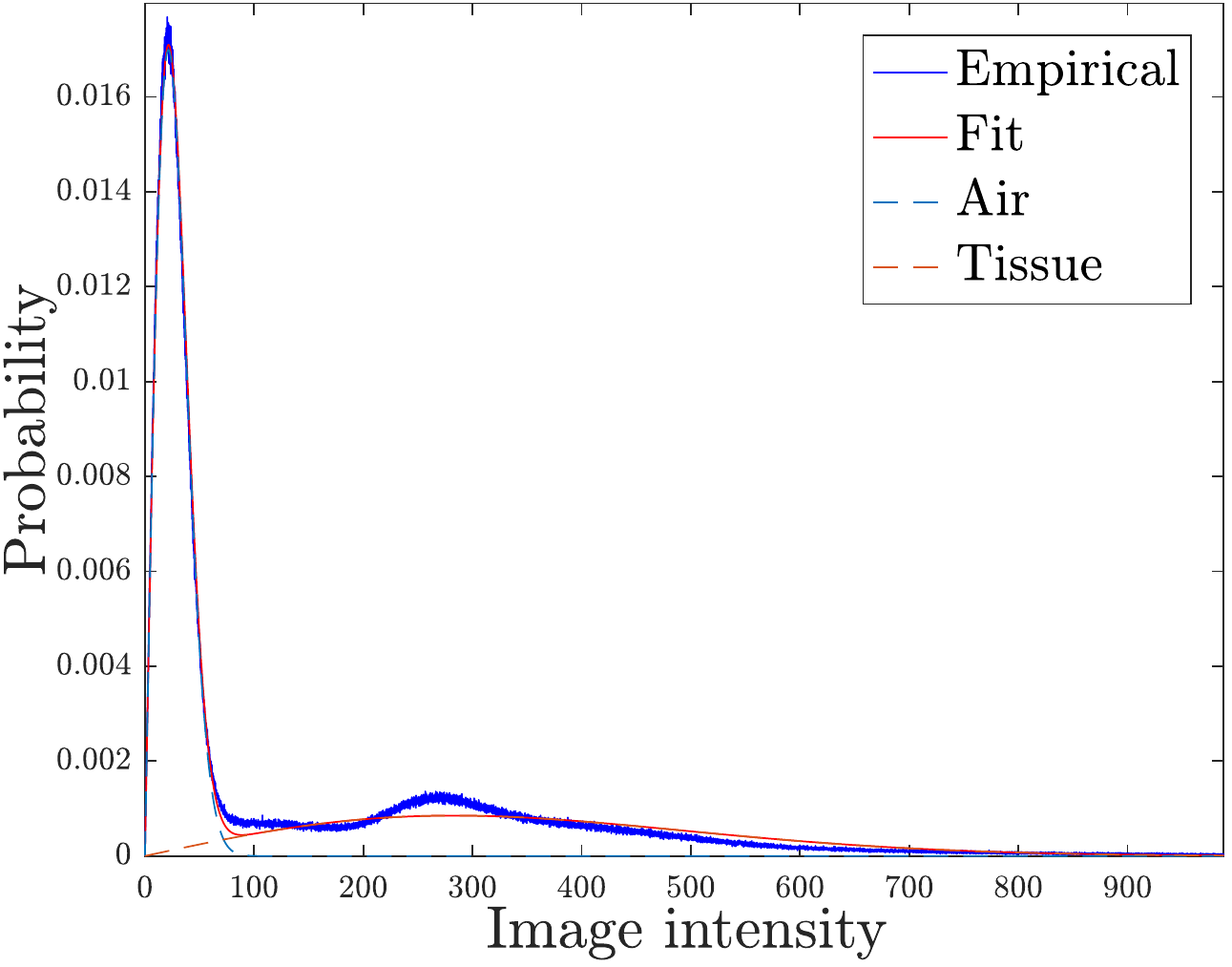}} \quad
	\subfloat[]{\includegraphics[width=0.31\textwidth]{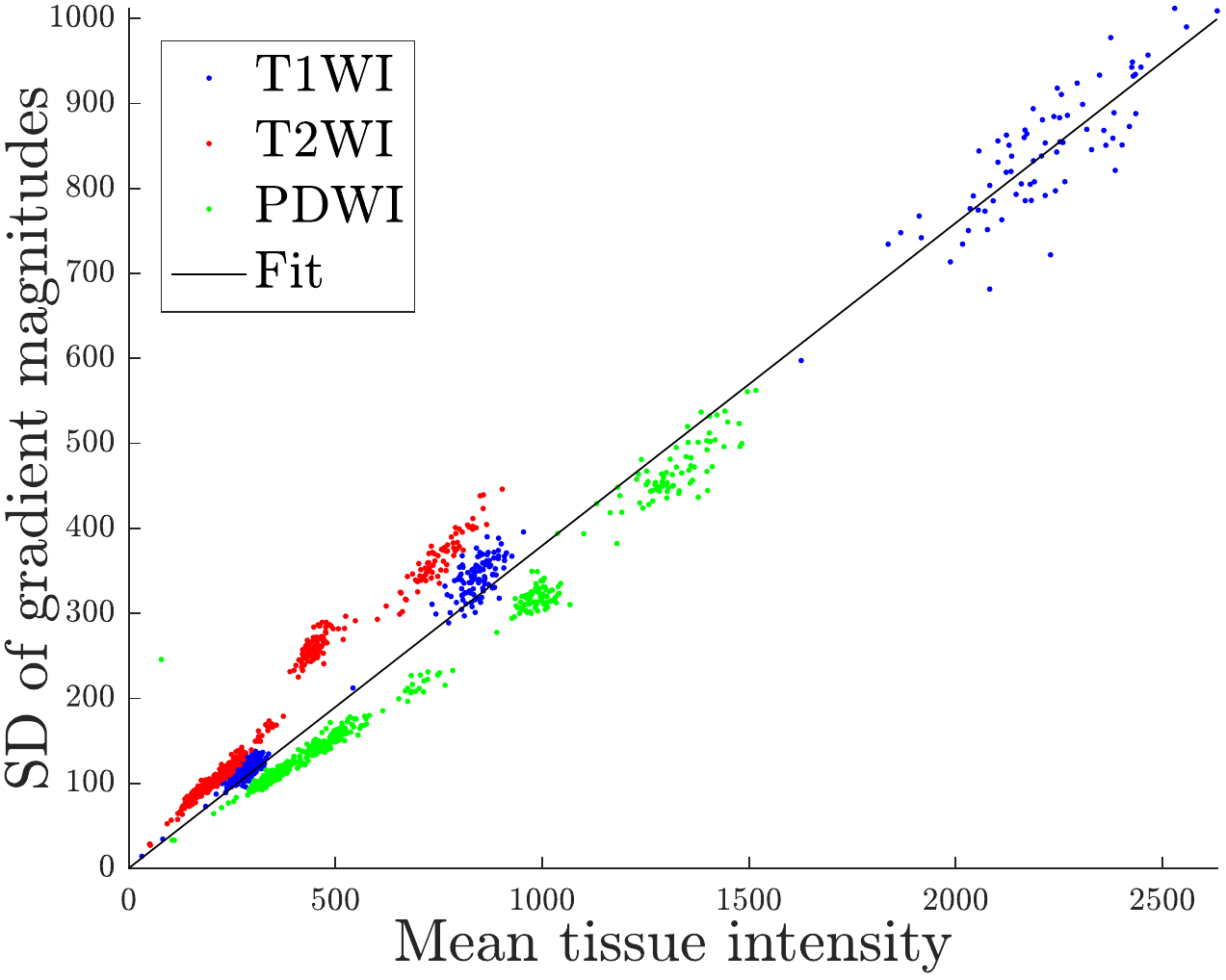}}	
	\caption{Estimating the hyper-parameters of the proposed SR model. a) Example of modelled slice profiles for increasing slice thicknesses. b) Mixture of two Rician distributions fitted to the intensity histogram of an MR scan. c) Least-squares fit between mean tissue intensities and standard deviation of gradient magnitudes.}
	\label{fig:improving:params}
\end{figure*}

\subsection{Setting of Hyper-Parameters}
For a generative model to generalise well to new observations, required hyper-parameters need to be either modelled as random variables or estimated from the observed data. Without a principled way of choosing these parameter, the model would not produce accurate results and a user would have to resort to manual tuning. Here, the hyper-parameters are the projection matrices, the noise precisions and the regularisation parameters.  This section gives more detail about how these parameters are chosen.

\begin{description}
	\item[Projection matrices] The projection matrices ${\bf A}_{ms}$ incorporates into the model an assumption of how each LR image was generated. The slice-selection is what generates the thick-sliced LR images and is therefore what should be modelled by these matrices. Mathematically, the slice selection can be expressed as the result of a sequence of linear operators on an unknown HR image, consisting of a geometrical transformation, application of the slice-profile and down-sampling. That is, the $s$th LR image, from the $m$th modality, is assumed drawn from the following observation model:
	\begin{equation}
	{\bf x}_{ms} = {\bf A}_{ms}{\bf y}_m + \bm{\eta}_{ms} = {\bf D}_{ms}{\bf S}_{ms}{\bf y}_m + \bm{\eta}_{ms},
	\label{eq:obs-model}
	\end{equation}
	where ${\bf S}$ simulates the slice-select profile of the MRI acquisition (see Figure \ref{fig:improving:params}a), ${\bf D}$ represents the down sampling of the HR image to the LR grid and $\bm{\eta}_{ms} \sim \mathcal{N}_{N_{ms}}(0,\tau_{ms}^{-1}{\bf I})$. Both ${\bf S}$ and ${\bf D}$ are dependent on a geometrical transformation ${\bf T} \in \mathbb{R}^{4 \times 4}$ that can be read from the image header. The observation model in \eqref{eq:obs-model} is well established for SR of MR scans \cite{greenspan2002mri,shilling2009super}.
	
	\item[Precision of the conditional distribution] The trade-off between fitting the data and keeping the result smooth will depend on how noisy each LR image is. Because MR scans are reconstructed as the magnitude of an image that was originally complex, the assumed Gaussian noise model is just an approximation of the true Rician noise model. It is therefore of interest to estimate the amount of Rician noise in each observed MR scan, because this would in turn give estimates of the parameters $\tau_{ms}$. In this work, a mixture of two Rician distributions is therefore fitted to the intensity histogram of each MR scan \cite{ashburner2013symmetric}, and the precision of the observation noise is calculated from the class corresponding to air (see Figure \ref{fig:improving:params}b).
	
	\item[Regularisation parameter] From a probabilistic modelling point of view, TV corresponds to the following Laplace distribution:
	\begin{align}
	p ({\bf y})&
	= Z^{-1} f ({\bf y}) = Z^{-1} \prod_{n=1}^{N} \exp \left( - \lambda \norm{ {\bf D}_n {\bf y} }_2 \right),
	\end{align}
	where $\lambda = 1/b$ and $b=\sqrt{\sigma^2 / 2}$. The hyper-parameter $\lambda$ can therefore be related to the standard deviation ($\sigma$) of the gradient magnitude ($\norm{{\bf D}_n {\bf y} }_2$) of the voxels in HR MR scans. This relationship is here used to obtain estimates of each $\lambda_m$ in \eqref{eq:improving:prior2}. More specifically, a least-squares fit is performed on the average tissue intensities, and the standard deviations of the gradient magnitudes, computed from a collection of 1,728 T1-, T2- and PD-weighted MR scans from the IXI dataset, resliced to have 1 mm isotropic voxel size (see Figure \ref{fig:improving:params}c). From this fit, a scaling function relating the image intensity of a LR image to the standard deviation of the gradient magnitude in a HR image can be constructed to give an estimate of a $\lambda_m$.
\end{description}

To summarise, a mapping for the regularisation parameters are learnt from HR images, the noise precisions are estimated from the LR data, and the projection matrices are based on metadata (slice thickness).

\begin{figure*}
	\centering
	\includegraphics[width=\textwidth]{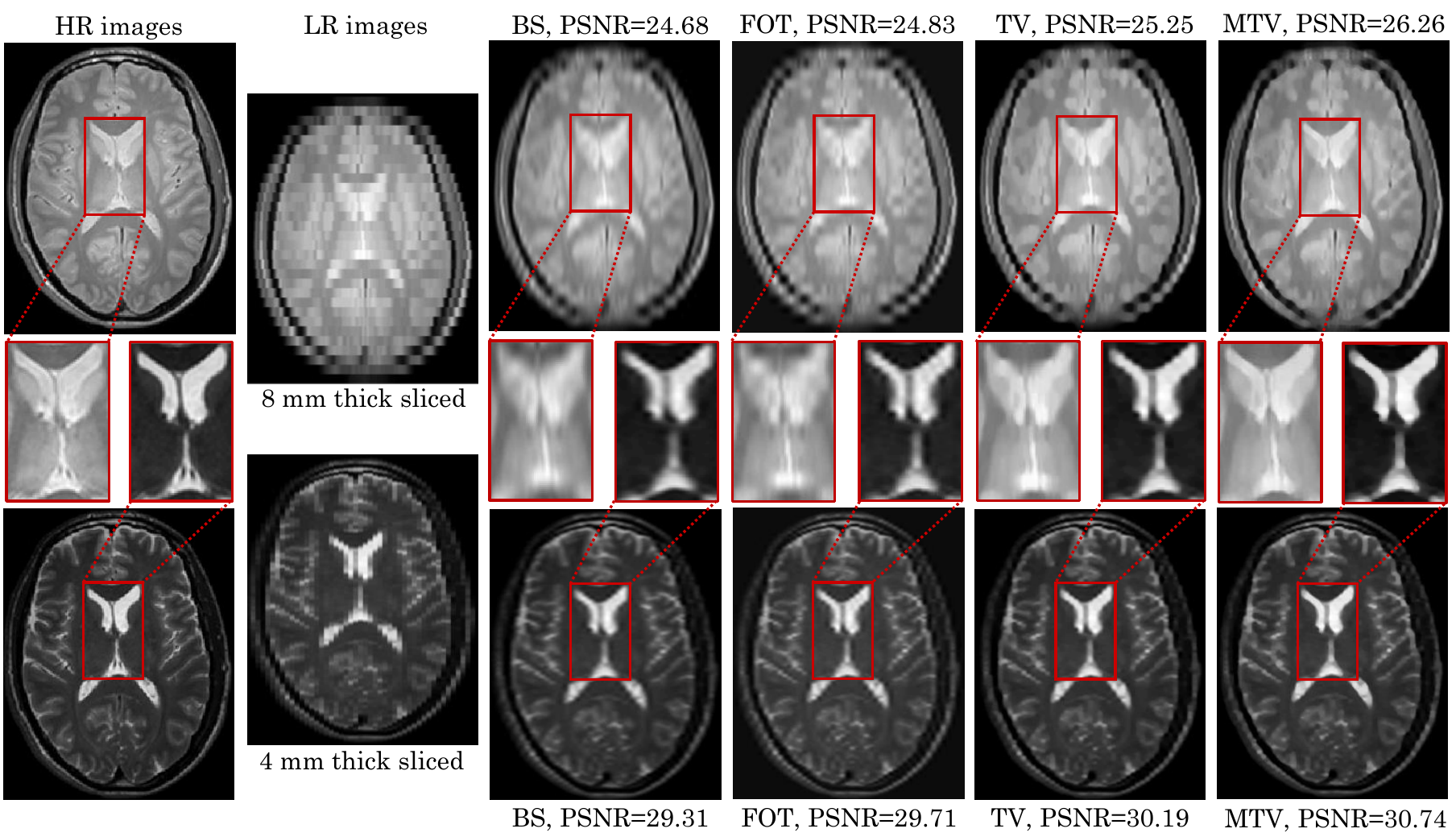}
	\caption{Two LR images were simulated from two HR images (PD- and T2-weighted) and then reconstructed using four methods. BS, FOT and TV are single-modality SR techniques, capable of combining LR images of only one MR contrast. MTV on the other hand, uses information from both MR contrasts.}
	\label{fig:improving:recs}
\end{figure*}

\section{Experiments and Results}
The proposed model was validated on the IXI dataset. LR images were generated from HR images allowing for ground-truth comparisons. The peak signal-to-noise ratio (PSNR), root-mean-square error (RMSE) and dice score were used to quantify the findings. Four methods for obtaining estimates of the unknown HR images were used. One interpolation based, using cubic b-splines (BS); three MAP based, using first-order Tikhonov (FOT), TV and MTV regularisation. FOT regularisation was achieved by changing the prior in \eqref{eq:improving:prior} accordingly. Figure \ref{fig:improving:recs} shows an example of simulating LR images from HR images, and subsequently estimating the HR images from the LR ones using the different methods. 

\begin{figure*}
	\centering 	
	\includegraphics[width=0.5\textwidth]{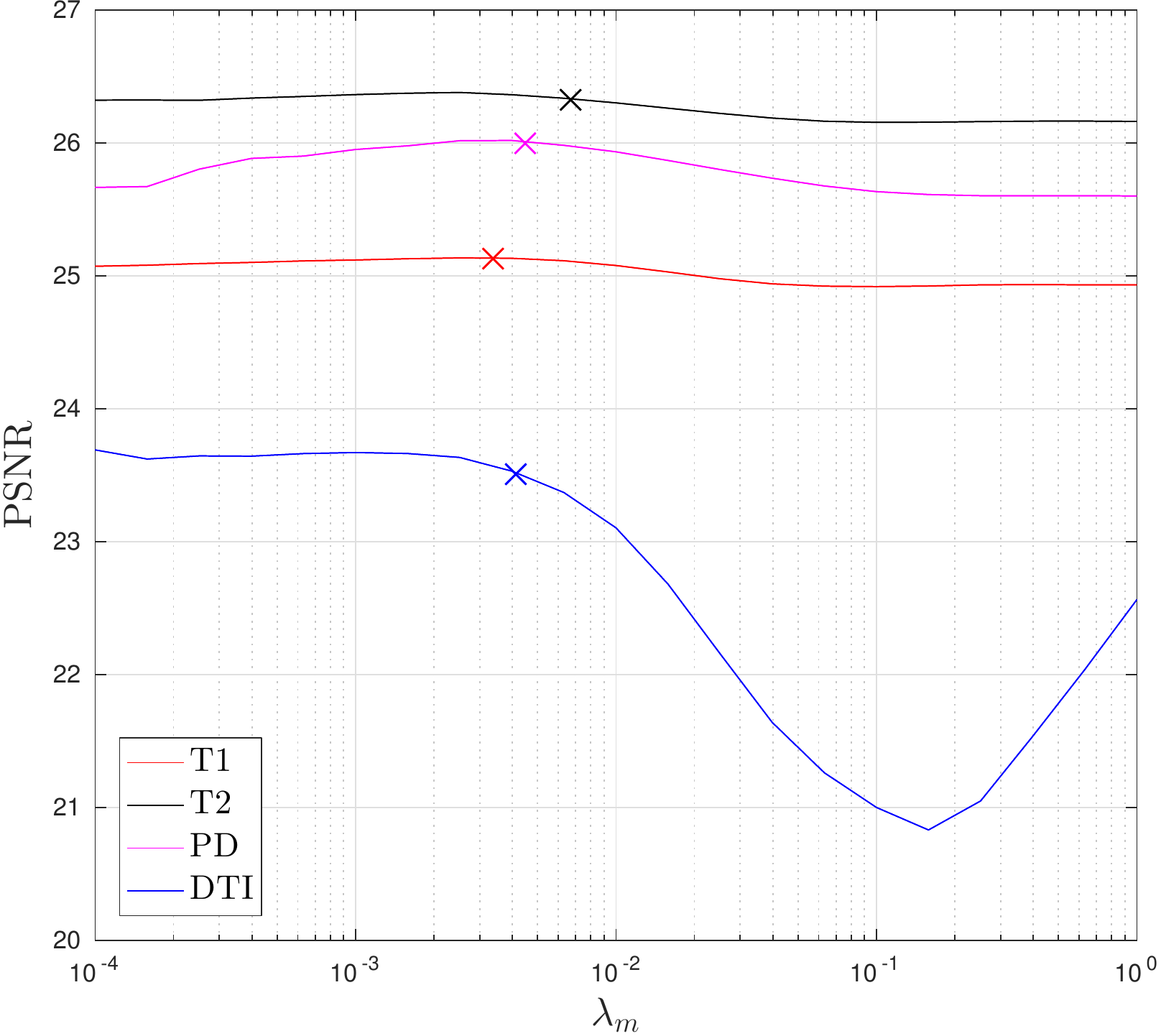}
	\caption{Grid-search over the regularisation parameters $\lambda_m$, for four different MR contrasts. The estimates of the regularisation parameterers are marked by crosses.}
	\label{fig:est-lam}
\end{figure*}

\begin{figure*}
	\centering 	
	\subfloat{\includegraphics[height=4.0cm,width=0.47\textwidth]{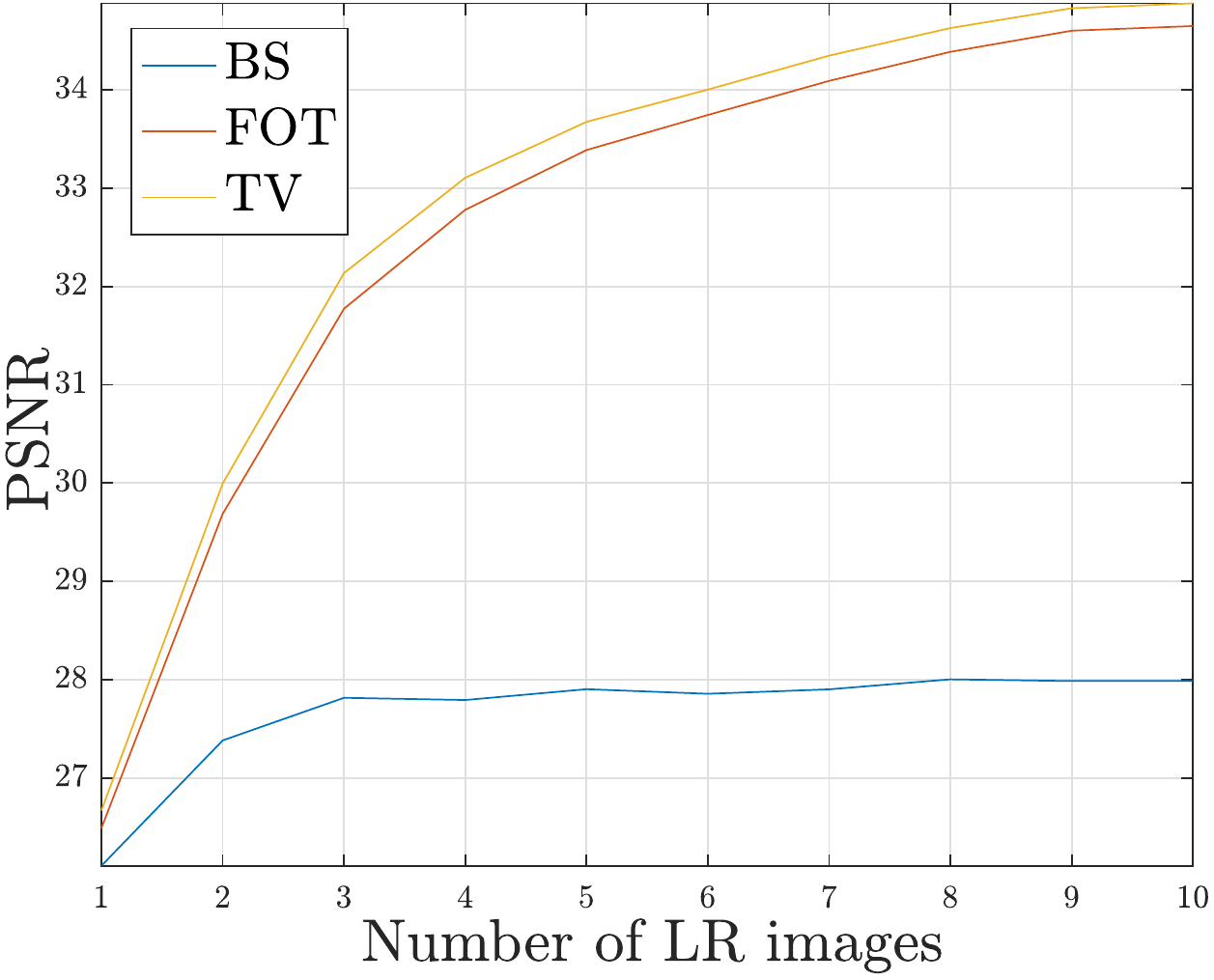}} \quad
	\subfloat{\includegraphics[height=4.1cm,width=0.47\textwidth]{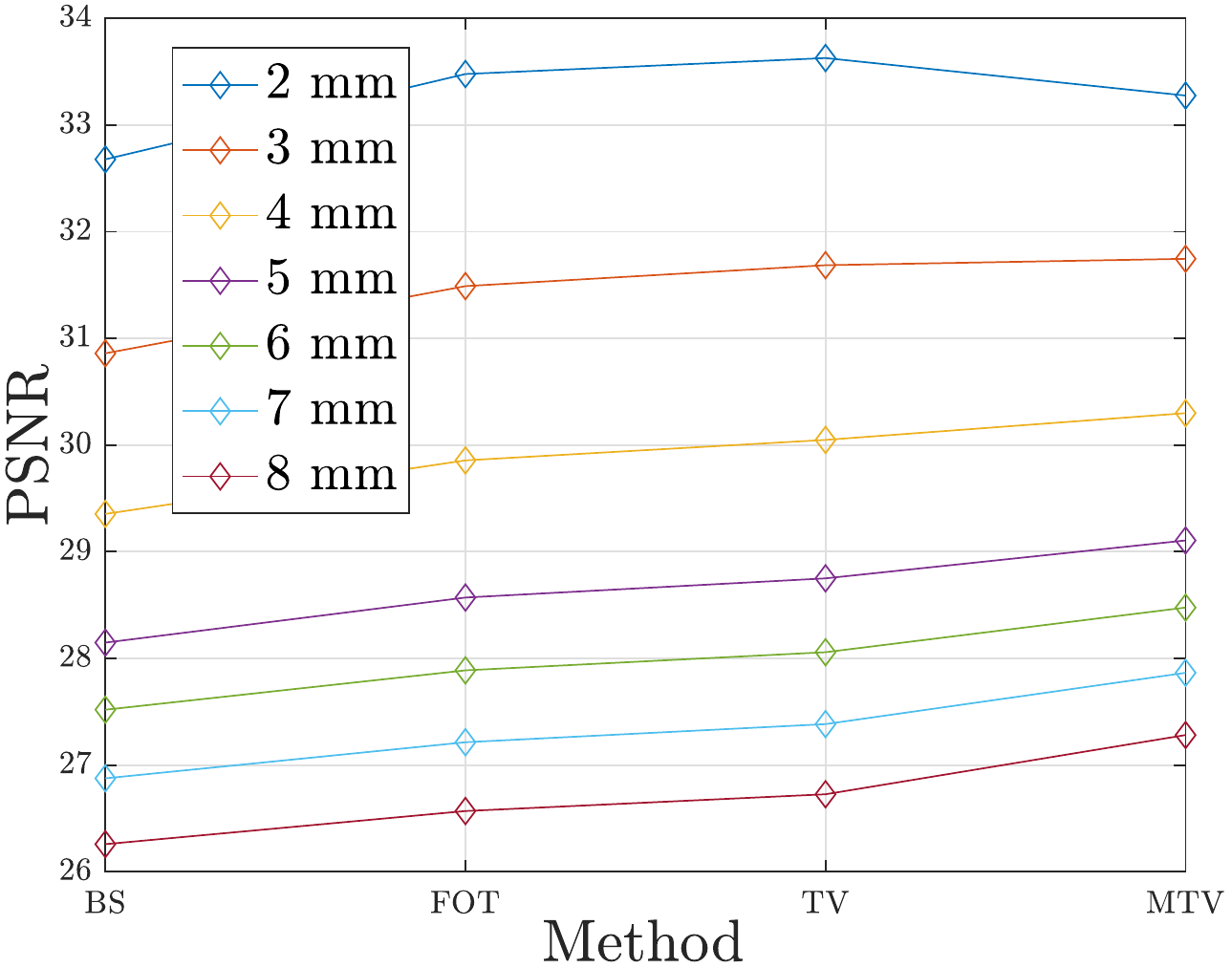}}
	\caption{a) PSNR for an increasing number of LR images. b) PSNR for increasing slice thicknesses.}
	\label{fig:plots}
\end{figure*}

The following five experiments were conducted:

\begin{description}
	\item[1. Estimating the observation noise] 1,728 images were used to verify the fit of the Rician mixture model. For each image, a known percentage of Rician noise was added (1\%, 2.5\%, 5\%, 10\%). The Rician mixture model was then fitted to the intensity histogram of the noisy image. From the fit, an estimate of the added noise was obtained. Table \ref{tab:improving:noise_estimate} shows the results of this estimation. From these results it is clear that the Rician mixture model accurately estimates a wide range of noise levels.
	
	\item[2. Estimating the regularisation parameter] Four MR scans, from different subjects and of different contrasts (T1-, T2-, PD- and DTI-weighted), were used to validate whether the estimate of the regularisation parameter $\lambda_m$ resulted in accurate reconstructions. The LR images were simulated having slice-thickness seven times greater than the in-plane resolution. Grid-searches over the regularisation parameters were then performed. Each grid-search was done in the range 10e\{-4:0.2:1\}. For each value of the regularisation parameter, the PSNR was computed between the resulting reconstruction and the known ground-truth. The DTI-weighted image was included to enable investigating if the method generalises to MR contrasts not part of the training set. The result of each grid-search, with the corresponding estimates marked by crosses, can be seen in Figure \ref{fig:est-lam}. This shows that the method for estimating the regularisation parameter produces good reconstructions, even for unseen MR contrasts.
		
	\item[3. Increasing number of LR images] An experiment was conducted to verify that image quality improves for an increasing number of LR images. For 20 subjects, LR images were simulated with a slice thickness seven times greater than the in-plane resolution. Every three LR images were simulated having orthogonal thick slice directions. For more than three images, simulations additionally included a 1 mm translational shift. HR images were reconstructed using BS averaging, FOT and TV. Figure \ref{fig:plots}a shows the results of the experiment. It can be seen that the reconstruction quality increases as more LR images become available. Furthermore, TV consistently reconstructs the highest quality images. As for the optimal number of LR images, it seems as if four images would provide the best trade-off between acquisition time and reconstruction quality. 
	
	\item[4. Multi-modality SR] The subjects of the IXI dataset having T1-, T2- and PD-weigted images were used (576 subjects) to quantify the benefit of reconstructing from subjects imaged with multiple MR contrasts. For each subject's T1-, T2- and PD-weighted data, LR images were simulated by picking the thick-slice direction and the downsampling factor at random (varying from a factor of two to eight). HR images were then reconstructed using BS, FOT, TV and MTV. Table \ref{tab:improving:psnr} shows the average PSNR where MTV reconstructed images with the greatest mean PSNRs and lowest standard deviation (sd). Figure \ref{fig:plots}b additionally shows average PSNR for different slice thicknesses, in which MTV once again performs favourably. However, for small slice thickness (2 mm) MTV was outperformed by both FOT and TV. This could be due to inaccurate estimates of the noise precisions and/or regularisation parameters, or misregistration of the LR images.
	
	\item[5. Predicting age from brain segmentations] To assess the value of SR in segmentation tasks, all the reconstructions from Experiment 4, as well as the HR references, were segmented into grey matter, white matter and cerebrospinal fluid using the multi-channel segmentation algorithm in the SPM software\footnote{Available from \url{http://www.fil.ion.ucl.ac.uk/spm/}.}. Dice scores were then calculated between the HR reference segmentation and the segmentations obtained from the reconstructions. The results can be seen in  Table \ref{tab:improving:ds} where MTV obtained the highest dice score. All of the segmentations were also given as training data, together with age labels, to a support vector machine regressor from which age predictions were performed. Table \ref{tab:improving:pred} shows RMSEs in predicting age. Interestingly, the RMSE in age from training using the MTV reconstructed segmentations was actually smaller than the RMSE from using the HR reference data. This result could be due to smoothing introduced in the reconstructed data.
\end{description}

\begin{table}
	\fontsize{7}{7.2}\selectfont
	\centering
	\caption{Estimates of Gaussian noise precisions from 1,728 IXI images.}
	\begin{widetable}{\columnwidth}{*{6}{c}} \toprule
		Ground-truth (\%)&0&1&2.5&5&10\\\midrule
		Estimate (\%)& $0.46\pm0.49$ & $1.12\pm0.78$ & $2.66\pm0.71$ &$5.00\pm1.02$ &$9.67\pm3.52$ \\\bottomrule
	\end{widetable}
	\label{tab:improving:noise_estimate}
\end{table}

\begin{table}
	\fontsize{7}{7.2}\selectfont
	\centering
	\caption{PSNR for different reconstruction methods. Shown as mean$\pm$std.}
	\begin{widetable}{\columnwidth}{*{5}{c}} \toprule
		Modality & BS & FOT & TV & MTV \\\midrule 
		T1& $29.83 \pm 9.42$ & $ 30.41 \pm 10.70$ & $30.62 \pm 10.81$& $30.73 \pm 9.35$
		\\\midrule  
		T2 & $28.16 \pm 8.15$ & $28.63 \pm 8.92$& $28.76 \pm 8.94$& $29.02 \pm 7.75$
		\\\midrule  
		PD & $28.46 \pm 9.97$ & $28.84 \pm 10.07$& $29.02 \pm 9.92$& $29.42 \pm 8.46$
		\\\bottomrule
	\end{widetable}
	\label{tab:improving:psnr}
\end{table}

\begin{table}
	\fontsize{7}{7.2}\selectfont
	\centering
	\caption{Dice scores for different reconstruction methods. Shown as mean$\pm$sd.}
	\begin{widetable}{\columnwidth}{*{5}{c}} \toprule
		Class & BS & FOT & TV & MTV \\\midrule 
		GM& $0.831 \pm 0.002$ & $ 0.839 \pm 0.002$& $0.836 \pm 0.002$& $0.848 \pm 0.002$
		\\\midrule  
		WM & $0.867 \pm 0.002$ & $0.874 \pm 0.001$& $0.871 \pm 0.001$& $0.881 \pm 0.001$
		\\\midrule  
		CSF & $0.852 \pm 0.001$ & $0.858 \pm 0.001$& $0.855 \pm 0.001$& $0.866 \pm 0.001$
		\\\bottomrule
	\end{widetable}
	\label{tab:improving:ds}
\end{table}

\begin{table}
	\fontsize{7}{7.2}\selectfont
	\centering
	\caption{RMSE for predicting age using 558 multi-modality brain segmentations. Shown as mean$\pm$sd.}
	\begin{widetable}{\columnwidth}{*{6}{c}} \toprule
		Predicted & HR & BS & FOT & TV & MTV \\\midrule
		RMSE (years)& $6.95 \pm 0.96$ & $7.00 \pm 1.24$ & $6.85 \pm 1.27$ & $6.71 \pm 1.23$ & $6.68 \pm 1.21$
		\\\bottomrule
	\end{widetable}
	\label{tab:improving:pred}
\end{table}

\section{Discussion}
This paper presented a generative model for SR in routine clinical MRI. A validation on a large number of subjects demonstrated that the model can recover information from a collection of LR subject MR scans. The MTV norm was introduced as a prior for SR in MRI and was shown able to recover information between MR scans of different contrasts. The model can be used to recover resolution on any set of patient MR scans, of arbitrary MR contrast and orientation, with good results. This is because the necessary hyper-parameters are estimated from the observed data, eliminating any need for parameter tuning in order to obtain accurate HR reconstructions. As a lack of generalisably is a common drawback among published algorithms, this is a strength of the proposed model. Furthermore, as reconstruction results improve with the availability of more LR images (see Figure \ref{fig:plots}a), the model should generalise well to patient data, in which anatomical information is distributed over multiple LR scans.   

The model could be of value in translating methods that have shown good results on research data to clinical imaging. For example, many techniques based on machine learning show promising results on analysing neuroimaging data. However, the amount of available high quality training images is limited, since it is both time-consuming and expensive to acquire large quantities of HR scans. As hospitals contain huge amounts of population representative data, accumulated over years of clinical service and available for research at cost neutrality, resolution recovery on this data could provide these algorithms with an abundance of training material. Furthermore, the proposed model also has the potential of improving the results of multi-channel segmentation algorithms, in which interpolation schemes usually are necessary to reslice images to the same size. Here, the model has the potential of decreasing partial volume effects and to create more accurate resized images, which could improve the segmentation output.

Applying SR to clinical MRI is a complex problem and unavoidably leads to limitations in the proposed model. For example, MRI scanner parameters, such as slice profile and slice gap, are highly variable and assuming such parameters as fixed can lead to inaccuracies. However, it should be possible to instead estimate these parameters from the data, by including them in the optimisation procedure. Misalignment between scans could be dealt with in a similar manner, rather than by performing an initial rigid registration of the LR images, as is currently done. A well known fact in image SR is that TV introduces stair-casing effects on flat areas, which are abundant in brain MRI. Using other types of regularisation have shown effective in decreasing such artefacts \cite{odille2015motion}. However, the MTV regularisation proposed in this paper seems not to suffer from stair-casing artefacts. This is probably due to the fact that MTV penalises based on gradients distributed over MR scans with often, orthogonal thick-slice directions. Hence, an image area containing flat gradients in one MR contrast may very well have more informative gradients in another contrast. Finally, the model will need to be validated on real patient data, which should be done in cooperation with an expert physician. This validation will be part of future work.

\bibliography{bibliography}

\begin{thebibliography}{10}

\bibitem{roobottom2010radiation}
C.~Roobottom, G.~Mitchell, and G.~Morgan-Hughes, ``{Radiation-reduction
  strategies in cardiac computed tomographic angiography},'' {\em Clin Radiol},
  vol.~65, no.~11, pp.~859--867, 2010.

\bibitem{Smith2018263}
S.~M. Smith and T.~E. Nichols, ``Statistical challenges in “big data” human
  neuroimaging,'' {\em Neuron}, vol.~97, no.~2, pp.~263 -- 268, 2018.

\bibitem{blaiotta2017generative}
C.~Blaiotta, P.~Freund, M.~J. Cardoso, {\em et~al.}, ``{Generative
  diffeomorphic modelling of large MRI data sets for probabilistic template
  construction},'' {\em NeuroImage}, 2017.

\bibitem{havaei2017brain}
M.~Havaei, A.~Davy, D.~Warde-Farley, A.~Biard, {\em et~al.}, ``{Brain tumor
  segmentation with deep neural networks},'' {\em Med Image Anal}, vol.~35,
  pp.~18--31, 2017.

\bibitem{ronneberger2015u}
O.~Ronneberger, P.~Fischer, and T.~Brox, ``{U-net: Convolutional networks for
  biomedical image segmentation},'' in {\em MICCAI}, pp.~234--241, Springer,
  2015.

\bibitem{van2012super}
E.~Van~Reeth, I.~W. Tham, C.~H. Tan, {\em et~al.}, ``{Super-resolution in
  magnetic resonance imaging: A review},'' {\em Concepts Magn Reson}, vol.~40,
  no.~6, pp.~306--325, 2012.

\bibitem{plenge2012super}
E.~Plenge, D.~H. Poot, M.~Bernsen, G.~Kotek, {\em et~al.}, ``{Super-resolution
  methods in MRI: Can they improve the trade-off between resolution,
  signal-to-noise ratio, and acquisition time?},'' {\em Magn Reson Med},
  vol.~68, no.~6, pp.~1983--1993, 2012.

\bibitem{greenspan2002mri}
H.~Greenspan, G.~Oz, N.~Kiryati, {\em et~al.}, ``{MRI inter-slice
  reconstruction using super-resolution},'' {\em Magn Reson Imag}, vol.~20,
  no.~5, pp.~437--446, 2002.

\bibitem{poot2010general}
D.~H. Poot, V.~Van~Meir, and J.~Sijbers, ``{General and efficient
  super-resolution method for multi-slice MRI},'' in {\em MICCAI},
  pp.~615--622, Springer, 2010.

\bibitem{manjon2010non}
J.~V. Manj{\'o}n, P.~Coup{\'e}, A.~Buades, {\em et~al.}, ``{Non-local MRI
  upsampling},'' {\em Med Image Anal}, vol.~14, no.~6, pp.~784--792, 2010.

\bibitem{alexander2014image}
D.~C. Alexander, D.~Zikic, J.~Zhang, {\em et~al.}, ``{Image quality transfer
  via random forest regression: applications in diffusion MRI},'' in {\em
  MICCAI}, pp.~225--232, Springer, 2014.

\bibitem{wang2016accelerating}
S.~Wang, Z.~Su, L.~Ying, {\em et~al.}, ``{Accelerating magnetic resonance
  imaging via deep learning},'' in {\em ISBI}, pp.~514--517, IEEE, 2016.

\bibitem{ebner2018volumetric}
M.~Ebner, K.~K. Chung, F.~Prados, {\em et~al.}, ``{Volumetric reconstruction
  from printed films: Enabling 30 year longitudinal analysis in MR
  neuroimaging},'' {\em NeuroImage}, vol.~165, pp.~238--250, 2018.

\bibitem{odille2015motion}
F.~Odille, A.~Bustin, B.~Chen, {\em et~al.}, ``{Motion-corrected,
  super-resolution reconstruction for high-resolution 3D cardiac cine MRI},''
  in {\em MICCAI}, pp.~435--442, Springer, 2015.

\bibitem{manjon2010mri}
J.~V. Manj{\'o}n, P.~Coup{\'e}, A.~Buades, {\em et~al.}, ``{MRI superresolution
  using self-similarity and image priors},'' {\em Int J Biomed Imag},
  vol.~2010, p.~17, 2010.

\bibitem{rousseau2010non}
F.~Rousseau, A.~D.~N. Initiative, {\em et~al.}, ``{A non-local approach for
  image super-resolution using intermodality priors},'' {\em Med Image Anal},
  vol.~14, no.~4, pp.~594--605, 2010.

\bibitem{jafari2014mri}
K.~Jafari-Khouzani, ``{MRI upsampling using feature-based nonlocal means
  approach},'' {\em IEEE Trans Med Imag}, vol.~33, no.~10, pp.~1969--1985,
  2014.

\bibitem{wen2008efficient}
Y.-W. Wen, M.~K. Ng, and Y.-M. Huang, ``{Efficient total variation minimization
  methods for color image restoration},'' {\em IEEE Trans Image Process},
  vol.~17, no.~11, pp.~2081--2088, 2008.

\bibitem{boyd2011distributed}
S.~Boyd, N.~Parikh, E.~Chu, {\em et~al.}, ``{Distributed optimization and
  statistical learning via the alternating direction method of multipliers},''
  {\em Found Trends Mach Learning}, vol.~3, no.~1, pp.~1--122, 2011.

\bibitem{dohmatob2014benchmarking}
E.~D. Dohmatob, A.~Gramfort, B.~Thirion, {\em et~al.}, ``{Benchmarking solvers
  for TV-ℓ 1 least-squares and logistic regression in brain imaging},'' in
  {\em PRNI}, pp.~1--4, IEEE, 2014.

\bibitem{shilling2009super}
R.~Z. Shilling, T.~Q. Robbie, T.~Bailloeul, {\em et~al.}, ``{A super-resolution
  framework for 3-D high-resolution and high-contrast imaging using 2-D
  multislice MRI},'' {\em IEEE Trans Med Imag}, vol.~28, no.~5, pp.~633--644,
  2009.

\bibitem{ashburner2013symmetric}
J.~Ashburner and G.~R. Ridgway, ``{Symmetric diffeomorphic modeling of
  longitudinal structural MRI},'' {\em Front Neurosci}, vol.~6, p.~197, 2013.

\end{thebibliography}
\bibliographystyle{ieeetr}

\end{document}